# Discovering linguistic (ir)regularities in word embeddings through max-margin separating hyperplanes

**Noel Kennedy, Imogen Schofield, Dave C. Brodbelt, David B. Church, Dan G. O'Neill**
The Royal Veterinary College,
Hawkshead Lane, North Mymms, Hatfield, Herts, AL9 7TA, UK
nkennedy@rvc.ac.uk, ischofield6@rvc.ac.uk, dbrodbelt@rvc.ac.uk, dchurch@rvc.ac.uk,
doneill@rvc.ac.uk

## Abstract

We experiment with new methods for learning how related words are positioned relative to each other in word embedding spaces. Previous approaches learned constant vector offsets: vectors that point from source tokens to target tokens with an assumption that these offsets were parallel to each other. We show that the offsets between related tokens are closer to orthogonal than parallel, and that they have low cosine similarities. We proceed by making a different assumption; target tokens are linearly separable from source and unlabeled tokens. We show that a max-margin hyperplane can separate target tokens and that vectors orthogonal to this hyperplane represent the relationship between source and targets. We find that this representation of the relationship obtains the best results in discovering linguistic regularities. We experiment with vector space models trained by a variety of algorithms (Word2vec: CBOW/skip-gram, fastText, or GloVe), and various word context choices such as linear word-order, syntax dependency grammars, and with and without knowledge of word position. These experiments show that our model, SVMCos, is robust to a range of experimental choices when training word embeddings.

## 1 Introduction

Mikolov et al., 2013c first observed that it was possible to perform algebra with the vector representations of word embeddings: `king−man+woman≈queen`. This showed that word embeddings encoded syntactic and semantic properties even though they were not explicitly trained to perform this task. The early approaches used the analogical method to analyse these properties: `a` is to `b` as `c` is to `d`. These linguistic regularities were thought to be encoded as constant offset vector in the embedding space (Mikolov et al., 2013c) such that the offset vector which pointed from `a` to `b` was approximately the same as the offset vector which pointed from `c` to `d`. Various methods for calculating these offset vectors have been proposed (Levy and Goldberg, 2014b; Linzen, 2016). However, analysis showed that these approaches to calculating offsets were confounded by the fact that the target words `b` & `d` were often the closest tokens in the embedding space to their respective source tokens `a` & `c` (Linzen, 2016; Rogers et al., 2017); high accuracy was achievable on some tasks by ignoring the example 'a is to b' and simply selecting `d` as the token with the highest cosine similarity with `c` (Linzen, 2016). Indeed, the current best performing method for discovering linguistic regularities, LRCos, doesn't learn from offsets at all and takes a different approach:

*"what is related to France as Tokyo is related to Japan?" can be reformulated as "what word belongs to the same class as Tokyo and is the closest to France?"* (Drozd et al., 2016)

We find that the offsets themselves are remarkably irregular sharing a mean cosine similarity which is comparable to randomly sampled tokens. Despite the apparently diverse nature of the offsets, we show that they do encode linguistic regularities and show how to discover them. Our method, SVMCos, doesn't assume that vector offsets between source and target tokens are parallel, or close to parallel, but instead we make a weaker assumption that source and target tokens are linearly separable in the vector space. We fit a max-margin hyperplane between source and target tokens and find that vectors orthogonal to the hyperplane show state of the art performance in predicting token relatedness. We analyse the predictions made by our model and show that it works, not by accurately predicting where related tokens are in the vector space, but by finding directions which increase the cosine similarity of target tokens whilst reducing the cosine similarity of source and unlabelled tokens.

We define terms used throughout this paper as follows:

- **Vector space model (VSM)**: this a set of word embeddings which represent words (tokens) as a vector in a low dimensional vector space. Previous research has led to several methods for learning word embeddings and we focus on four popular algorithms: fastText (Bojanowski et al., 2017), GloVe (Pennington et al., 2014), and word2vec's skip-gram and continuous bag-of-words (Mikolov et al., 2013a).
- **Task**: The problem we address is evaluating how well different models perform at learning how a particular VSM encodes relations between tokens. BATS provided several categories of relations containing 50 examples for each category of paired source and target tokens. The source/target pairs share a relation. For example, the E01 category lists source countries paired with target capitals.
- **Source** and **target** tokens are words represented by vectors in the VSM which have a syntactic or semantic relationship (defined by the BATS dataset) e.g. `France/Paris, Spain/Madrid`
- **Offset(s)**: This is a vector in a VSM which is the vector difference between a source and target token. In our experimental design offsets point from a source token to the target token.
- **Models**: The models are trained such that: given a source token, a model must predict the correct target(s) which is equivalent to predicting the offset.
- **Query point**: The point in the VSM where a model predicts a target token will be for a given source token. Model query points are highly unlikely to land exactly on target tokens, or indeed any token, as VSMs are sparsely populated. Therefore, we search around the query point for tokens in the VSM that have the highest cosine similarity with the query point. A perfect model's query points would be equal to the offset added to the source token (in the case of multiple target tokens, this is impossible unless the target tokens are all represented by the same vector in the VSM).
- **Target word classification**: It turns out that there are usually tokens with high cosine similarity with query points which are not the expected target token. Therefore, it can sometimes benefit model accuracy these false positive tokens are given a lower ranking than the target tokens. We follow the method developed by Drozd et al., 2016 who trained a logistic regression (LR) classifier to separate target class tokens from source tokens and out-of-category tokens. For example, given a paired list of source/target tokens of capitals, `France/Paris, UK/London, Spain/Madrid`, the LR would learn to classify the target capitals, `Paris, London, Madrid`, as the positive class. Contrasting negative class examples were randomly sampled from the vocabulary. We describe which models which use this method below and when we experimented with model variants, we indicate this in the model name: **+LR** or **-LR**.

## 2 Method

Our experimental setup required models to produce a query point for each source token. If a model did not use target class classification, its score for a token in the VSM vocabulary was the cosine similarity of its query point with that token. If a model did use target class classification, its cosine score was multiplied by the LR's classification score. We ranked tokens by the models' score and calculated metrics against the resultant ranking and the gold standard provided by BATS.

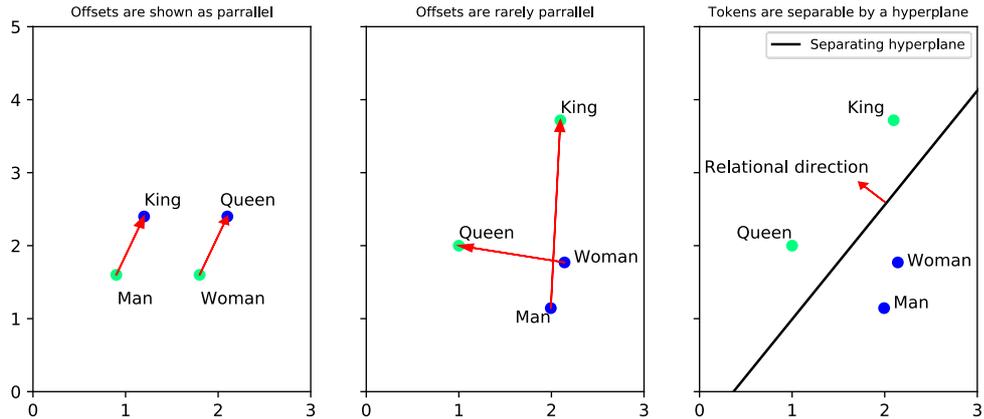

Figure 1 Relational offsets between related source/target tokens (`Woman`/`Queen` and `Man`/`King`) are shown. In the literature these offsets are sometimes shown as parallel (left chart) but we observed the mean cosine similarity of offsets to be ≤ 0.11. The middle chart shows two relational offsets with a more empirically based cosine similarity of 0.16. Our model, SVMCos, finds a relational direction from source tokens to target tokens (right chart) by fitting a max-margin hyperplane to separate source/target tokens. Vectors orthogonal to the hyperplane encode the relationship between source and target.

## 2.1 Models

All the models in this paper use query point prediction and some use target word classification as well. Some models (SVMCos, PCA+LR, and the neural networks) had hyperparameters which required tuning. Because of the risk of over-fitting with our experimental set-up, we chose the models' hyperparameters based on maximising the F1 score on a held-out dataset and two held-out VSMs.
**Our models**:
  **SVMCos**: Our model worked by fitting a separating hyperplane between source and target tokens in the VSM. Vectors orthogonal to the hyperplane pointed from source tokens to target (see Figure 1). The hyperplane gave us a direction, but not a magnitude. SVMCos chose a magnitude equal to the $\alpha$ percentile from the distribution of the Euclidean distances between source and target tokens observed in the training set. We found we could increase the performance of SVMCos by including samples of 20 unlabelled tokens nearest to the source token (nearest by cosine similarity). These unlabelled tokens were given the same class as source tokens. We used a linear support vector machine (SVM) (Cortes and Vapnik, 1995) to fit a max-margin hyperplane using the implementation in scikit-learn version 0.21.2 (Pedregosa et al., 2011). SVMCos performed target class classification. SVMCos's hyperparameters were set as follows: C=0.001, $\alpha$=25. As C was less than 1, this meant that the SVM learned hyperplanes with larger margins (it was allowed to misclassify tokens nearest the hyperplane). The value chosen for the $\alpha$ parameter meant that the magnitude of the relational vector was the size of the 25th percentile of the observed offset magnitudes. This meant that SVMCos's query points were closer to the source token than the observed target tokens.

  **PCA+LR:** We experimented with learning relational offset through principal component analysis (PCA). The relational offset was learned by taking the first principal component of the offset vectors seen in training. The magnitude chosen was the $25^{th}$ percentile of the observed magnitudes. PCA+LR performed target class classification.

  **NN Linear** & **NN non-linear**: We experimented with two neural networks which were trained using a regression task to produce query points: a linear neural network (no activation function) and a non-linear neural network with a single hidden layer with a ReLU activation function (Nair and Hinton, 2010). The neural nets did not perform target class classification.
**Previously described models:**
  **LRCos** and **3CosAvg**: These were chosen as our baselines (Drozd et al., 2016). LRCos was the first model to use what we refer to target class classification. It worked by training a logistic regression classifier to identify members of the target. LRCos's query point was the embedding of the source token. Tokens were scored by multiplying these two factors. 3CosAvg's query points were made by simply adding the embedding of the source token to the mean of the offsets observed in training. 3CosAvg had no target class classification.

## 2.2 Dataset and metrics

We experimented with the Bigger Analogy Test Set (BATS) (Gladkova et al., 2016) which was curated to give insights into how VSMs capture general relations in text. BATS contained several categories of relations, including encyclopedic (e.g. country-capital `France:Paris`), synonyms (`afraid:terrified`), singular noun and their plurals, and derivational morphology (e.g. `aquired:reaquired`). BATS contained 99,200 pairs of related terms. The relational categories in BATS were balanced so each category contained 50 examples of each relationship.

For some relational categories, BATS specified multiple correct answers for a given source token. For example, in the name-occupation relation, the source `Darwin` had a number of truly related terms `naturalist`, `biologist` and `geologist`. We wanted to evaluate how many of these related terms a particular model could accurately predict. We chose two metrics to measure this; sensitivity and mean average precision. We calculated an F1 score using the harmonic mean of the sensitivity and mean average precision. Given the large number of combinations of models and VSMs in this paper, we chose to report only the F1 statistic and to report it at the dataset level, rather than at the relationship level. All results were micro-averaged.

## 2.3 Vector space models (VSMs)

Recall from the terminology defined previously that a vector space model (VSM) is a low dimensional space in which tokens are represented as points in the space (vectors). We identified several 19 pre-trained VSMs which were each trained on different corpora using a wide variety in training methods. We experimented using VSMs trained by four popular algorithms : fastText (Bojanowski et al., 2017), GloVe (Pennington et al., 2014) and word2vec's (Mikolov et al., 2013c); both skip-gram and continuous bag-of-words (CBOW) methods. Most of the 19 VSMs we experimented with (See Table 1) used the default word context of their respective algorithm but our 19 VSMs included the pre-trained VSMs of Li et al. (2017) who experimented with four different methods for representing word context. Previous research has shown that word embeddings are particularly sensitive to this context (Levy and Goldberg, 2014a; Rogers et al., 2018). Li et al.'s (2017) contexts were either linear or syntax dependency-based and bound or unbound (all four combinations of both pairs of choices were experimented with):

- bound : relative word order was preserved so if the word `football` was 2 words to the left of the target word it was relabelled `football/-2`. This meant, for example, `football/-2` could have a different representation in the VSM from `football/+2`
- unbound : word order was ignored so if the word `football` was 2 words to the left of the target word its label was `football`
- linear : the words next to the target word as defined by the word order of the sentence
- syntax dependency-based (deps) : a target word's syntactic neighbours as defined by a parser

See Table 1 for a full list of the 19 VSMs we experimented with, including model names and a short description of the training corpus and method. When a particular VSM lacked the vocabulary for a pair in the dataset, we dropped the pair from the evaluation.

Each relational category in BATS contained 50 examples of source tokens paired with target tokens. Our experimental setup allowed for multiple target tokens for a single source token. We trained our models with k-fold validation with k = 50 (leave-one-out). Our evaluation method was intrinsic.

All experiments were 'honest' (Drozd et al., 2016) in that there was no hardcoded rule which prevented models from predicting any of the source or target tokens seen in training. This is because target tokens tended to have high cosine similarity with their source tokens. Indeed, for some relations, it is possible to get accuracies as high as 70% by ignoring the training set and simply choosing the nearest embedding to the source token (Linzen, 2016).

| VSM Name | Description |
|---|---|
| fastText wiki | Wikipedia 2017, UMBC webbase corpus and statmt.org news dataset (16B tokens, 1M vocab)[1] (Mikolov et al., 2018) |
| fastText wiki subword | As above with sub-word information[1] |
| fastText CC | Common Crawl (60B tokens, 2M vocab)[1] |
| fastText CC subword | As above with sub-word information[1] |
| GloVe CC | Common Crawl (42B tokens, 1.9M vocab, uncased, 300d vector) (Pennington et al., 2014)[2] |
| GloVe wiki | Wikipedia 2014 + Gigaword 5 (6B tokens, 400K vocab, uncased, 300d vectors)[2] |
| Skip-gram news | Google news (100B tokens, 3M vocab)[3] (Mikolov et al., 2013b) |
| Li bound_deps CBOW | Wikipedia 2013 Li et al. (2017) Syntax-based dependencies and bound. CBOW[4] |
| Li bound_deps GloVe | As above but GloVe[4] |
| Li bound_deps skip-gram | As above but skip-gram[4] |
| Li bound_linear CBOW | Wikipedia 2013 Li et al. (2017) Linear-based dependencies and bound. CBOW[4] |
| Li bound_linear GloVe | As above but GloVe[4] |
| Li bound_linear skip-gram | As above but skip-gram[4] |
| Li unbound_deps CBOW | Wikipedia 2013 Li et al. (2017) Syntax-based dependencies and unbound. CBOW[4] |
| Li unbound_deps GloVe | As above but GloVe[4] |
| Li unbound_deps skip-gram | As above but skip-gram[4] |
| Li unbound_linear CBOW | Wikipedia 2013 Li et al. (2017) Linear-based dependencies and unbound. CBOW[4] |
| Li unbound_linear GloVe | As above but GloVe[4] |
| Li unbound_linear skip-gram | As above but skip-gram[4] |

Table 1: The nineteen vector space models (VSM) used in evaluation. We included a wide variety of pre-trained VSMs trained used one of four algorithms with a variety of word contexts and corpora. Please refer to original papers for further details such as hyperparameter choices. Where the creators of the VSMs provided multiple dimension sizes for the word embeddings, we used the largest.

## 3 Results

We found that the offsets between source and target tokens were much closer to orthogonal than parallel. The mean cosine similarity score of offsets (defined as vector difference of target-source) ranged from 0.05 (Li unbound_linear GloVe) to 0.11 (GloVe CC). Recall that vectors which point in the same direction have a cosine similarity of 1.0 and vectors at right angles to each other have a score of 0. next experimented by evaluating how the models would perform at finding related tokens, given the irregularity of their vector differences. SVMCos performed the best for all VSMs (see Table 2).

Recall that models could be categorized by two components: how the model created query points (where the model predicts target tokens to be in the vector space), and whether or not the model performed target class classification (to score the tokens near to the query point according to their similarity with the target class). We were interested in the relative contribution of the first factor: the models' ability to create accurate query points. This factor was critical in our experiments since all models that performed target class classification used the same method developed by Drozd et al., 2016: logistic regression trained on the same training set. In other words, the key performance differentiator of the models was the quality of its query points. We ran an experiment to compare only the models' query points by removing all target class classification from all models. The results are shown in Table 3. The results show that SVMCos's method for producing query points outperformed all other models in all VSMs.

---

[1] Downloaded from https://fasttext.cc/docs/en/english-vectors.html
[2] Downloaded from https://nlp.stanford.edu/projects/glove/
[3] Downloaded from https://code.google.com/archive/p/word2vec/
[4] Downloaded from https://vecto.readthedocs.io/en/docs/tutorial/getting_vectors.html

| Vector Space Model (See Table 1 for VSM definition) | 3CosAvg | LRCos | PCA +LR | SVMCos | NN Linear | NN Nonlinear |
|---|---|---|---|---|---|---|
| fastText CC | 0.44 | 0.65 | 0.51 | **0.68** | 0.15 | 0.19 |
| fastText CC subword | 0.46 | 0.55 | 0.38 | **0.59** | 0.14 | 0.16 |
| fastText wiki | 0.50 | 0.52 | 0.33 | **0.63** | 0.13 | 0.14 |
| fastText wiki subword | 0.50 | 0.52 | 0.33 | **0.63** | 0.14 | 0.13 |
| GloVe CC | 0.41 | 0.54 | 0.34 | **0.64** | 0.15 | 0.19 |
| GloVe wiki | 0.37 | 0.51 | 0.32 | **0.58** | 0.15 | 0.19 |
| Li bound_deps CBOW | 0.30 | 0.35 | 0.31 | **0.46** | 0.13 | 0.12 |
| Li bound_deps GloVe | 0.12 | 0.31 | 0.23 | **0.36** | 0.10 | 0.13 |
| Li bound_deps skip-gram | 0.32 | 0.38 | 0.29 | **0.53** | 0.14 | 0.12 |
| Li bound_linear CBOW | 0.31 | 0.40 | 0.35 | **0.52** | 0.14 | 0.14 |
| Li bound_linear GloVe | 0.23 | 0.36 | 0.27 | **0.46** | 0.15 | 0.18 |
| Li bound_linear skip-gram | 0.34 | 0.41 | 0.31 | **0.54** | 0.16 | 0.14 |
| Li unbound_deps CBOW | 0.34 | 0.45 | 0.31 | **0.51** | 0.16 | 0.16 |
| Li unbound_deps GloVe | 0.28 | 0.46 | 0.28 | **0.49** | 0.18 | 0.18 |
| Li unbound_deps skip-gram | 0.34 | 0.46 | 0.36 | **0.55** | 0.15 | 0.14 |
| Li unbound_linear CBOW | 0.33 | 0.48 | 0.38 | **0.52** | 0.15 | 0.15 |
| Li unbound_linear GloVe | 0.12 | 0.17 | 0.10 | **0.24** | 0.06 | 0.12 |
| Li unbound_linear skip-gram | 0.33 | 0.46 | 0.34 | **0.54** | 0.15 | 0.14 |
| Skip-gram news | 0.35 | 0.54 | 0.42 | **0.58** | 0.11 | 0.13 |

Table 2 Micro-average F1 scores on the BATS dataset. The models were evaluated on their ability to predict the related target tokens given a source token. Our model SVMCos obtained the best results by fitting a max-margin classifier to separate source vectors from target vectors and using the vector orthogonal to the separating hyperplane as the offset to locate related tokens in vector space.

Recall that models could be categorized by two components: how the model created query points (where the model predicts target tokens to be in the vector space), and whether or not the model performed target class classification (to score the tokens near to the query point according to their similarity with the target class). We were interested in the relative contribution of the first factor: the models' ability to create accurate query points. This factor was critical in our experiments since all models that performed target class classification used the same method developed by Drozd et al., 2016: logistic regression trained on the same training set. In other words, the key performance differentiator of the models was the quality of its query points. We ran an experiment to compare only the models' query points by removing all target class classification from all models. The results are shown in Table 3. The results show that SVMCos's method for producing query points outperformed all other models in all VSMs.

Given the higher performance of SVMCos, we wanted to investigate the nature of SVMCos's query points by comparing its query points with the query points of the previous state of the art model, LRCos. Recall that LRCos's query points are the embedding of the source token. In other words, LRCos did not make use of offsets and relied on the observation that target tokens had high cosine similarity with the source token. These results are charted in Figure 2.

## 4 Discussion

The best performance in predicting related tokens in all VSMs was SVMCos. LRCos performed second best in all VSMs. PCA+LR and 3CosAvg performed similarly in the main evaluation (Table 2) but it is clear from comparing the two models in Table 3 that PCA was much worse at producing query points and PCA+LR's relatively high performance in the main evaluation can be attributed to the logistic regression target classification component, not PCA's ability t. The two neural networks were particularly bad at this task. The non-linear network slightly outperformed the linear network.

Counter intuitively, we found that SVMCos obtained results superior to LRCos despite the fact that SVMCos's query points were actually further from the target tokens than LRCos's query points. As LRCos's query points were defined as the embedding of the source token, this meant, target tokens were actually closer to the source token than to SVMCos's query points in the embedding space. Figure 2 shows the following:
- Column 1: the Euclidean distance between source tokens and the target token were smaller for LRCos than for SVMCos. This meant that SVMCos's query points were actually further from the target tokens than if SVMCos had just assumed, as LRCos did, that source token were close to target tokens in the embedding space. However, Euclidean distance was not used by either LRCos or SVMCos to score tokens. Both used cosine similarity.
- Column 2: SVMCos's query points had higher cosine similarity with the target tokens than LRCos's despite lying further away in Euclidean distance.
- Column 3: SVMCos's query points had lower cosine similarity with non-target tokens (true negative) than LRCos's. These non-target tokens were sampled near the models' respective query points to produce these charts.

This is not shown on Figure 2 but we also compared the cosine similarity of both model's query points with respect to non-target tokens around the source token (as well as the source token itself). SVMCos's query points had lower cosine similarity with these true-negative tokens. This made sense as SVMCos's query points were further from the source token than LRCos's query points. This also explained why there was no need to make a 'dishonest' version of SVMCos as it rarely predicted source tokens in place of target tokens, a weakness of earlier models (Linzen, 2016; Rogers et al., 2017). The findings in column 1, 2 and 3 of Figure 2 show that SVMCos's superior performance in producing query points (Table 3) was obtained, not by predicting query points which accurately match the true offsets, but by learning directions in the hyperspace which separated target tokens from local source and non-target tokens.

| Vector Space Model (See Table 1 for VSM definition) | 3CosAvg | LRCos -LR | PCA -LR | SVMCos -LR | NN Linear | NN Nonlinear |
|---|---|---|---|---|---|---|
| fastText CC | 0.44 | 0.18 | 0.13 | **0.61** | 0.15 | 0.19 |
| fastText CC subword | 0.46 | 0.18 | 0.10 | **0.59** | 0.14 | 0.16 |
| fastText wiki | 0.50 | 0.20 | 0.10 | **0.61** | 0.13 | 0.14 |
| fastText wiki subword | 0.50 | 0.20 | 0.10 | **0.61** | 0.14 | 0.13 |
| GloVe CC | 0.41 | 0.20 | 0.14 | **0.55** | 0.15 | 0.19 |
| GloVe wiki | 0.37 | 0.20 | 0.13 | **0.50** | 0.15 | 0.19 |
| Li bound_deps CBOW | 0.30 | 0.14 | 0.13 | **0.34** | 0.13 | 0.12 |
| Li bound_deps GloVe | 0.12 | 0.04 | 0.05 | **0.15** | 0.10 | 0.13 |
| Li bound_deps skip-gram | 0.32 | 0.15 | 0.12 | **0.42** | 0.14 | 0.12 |
| Li bound_linear CBOW | 0.31 | 0.16 | 0.15 | **0.41** | 0.14 | 0.14 |
| Li bound_linear GloVe | 0.23 | 0.11 | 0.11 | **0.31** | 0.15 | 0.18 |
| Li bound_linear skip-gram | 0.34 | 0.15 | 0.12 | **0.47** | 0.16 | 0.14 |
| Li unbound_deps CBOW | 0.34 | 0.18 | 0.14 | **0.43** | 0.16 | 0.16 |
| Li unbound_deps GloVe | 0.28 | 0.15 | 0.09 | **0.36** | 0.18 | 0.18 |
| Li unbound_deps skip-gram | 0.34 | 0.18 | 0.14 | **0.48** | 0.15 | 0.14 |
| Li unbound_linear CBOW | 0.33 | 0.19 | 0.16 | **0.44** | 0.15 | 0.15 |
| Li unbound_linear GloVe | 0.12 | 0.07 | 0.03 | **0.15** | 0.06 | 0.12 |
| Li unbound_linear skip-gram | 0.33 | 0.17 | 0.12 | **0.47** | 0.15 | 0.14 |
| Skip-gram news | 0.35 | 0.20 | 0.15 | **0.51** | 0.11 | 0.13 |

Table 3 BATS micro-average F1 scores. We compare only the model's query points produced by the by removing the target class classification component from models which used it (indicated by -LR) LRCos-LR predicted that target tokens are nearest to the source token (no offset). These results show that fitting a max-margin separating hyperplane between source tokens and target tokens outperformed

all other methods for finding target tokens in the embedding space including the assumption that target tokens are closest to source tokens.

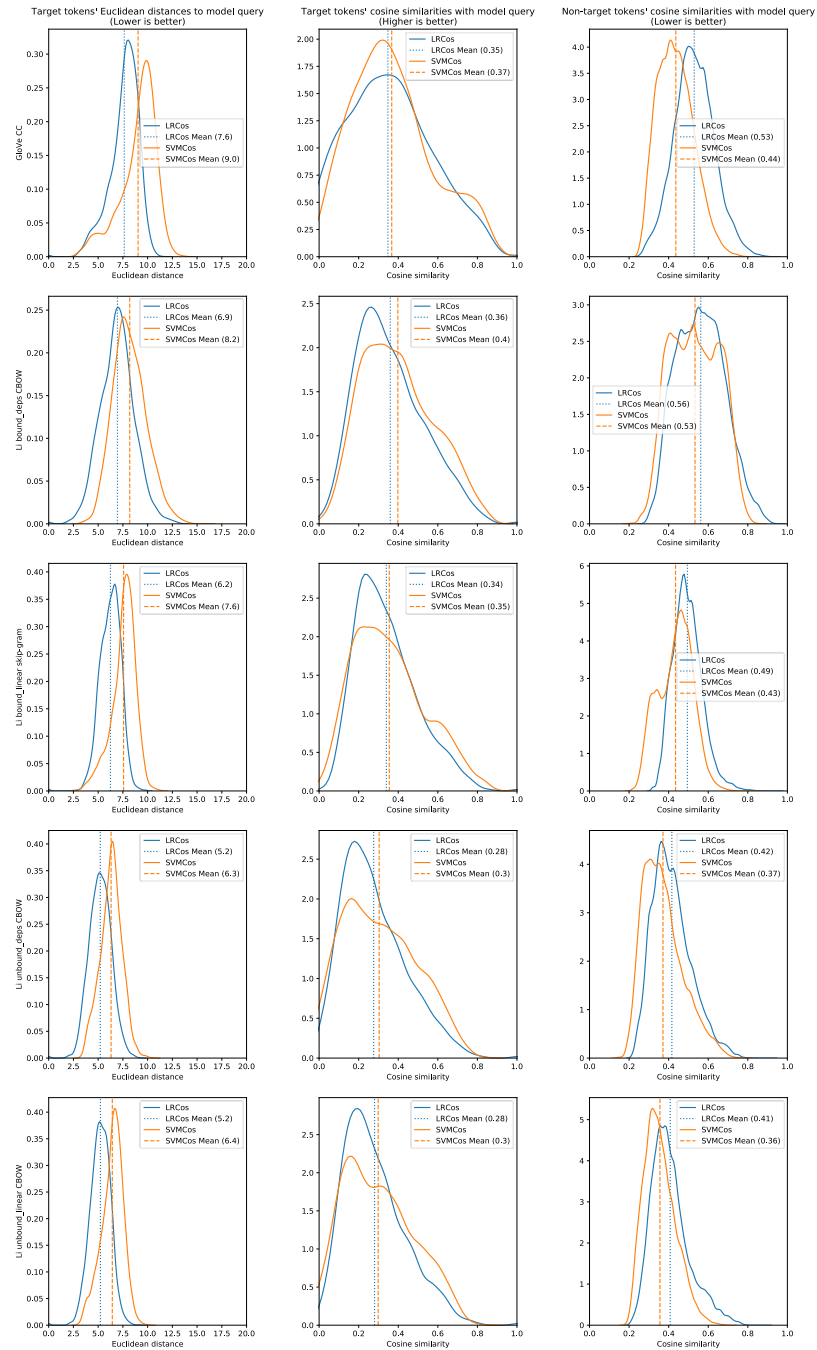

Figure 2 Kernel density estimates comparing the query points of SVMCos and LRCos. Each model, given a source token, predicts where in the vector space the related target tokens should be. SVMCos learns an offset in vector space to predict where target tokens are. LRCos assumes related tokens are near to the source token. Column 1 charts show that SVMCos's predictions are actually further from target tokens in Euclidean space than LRCos's predictions. However, column 2 shows SVMCos's predictions have a higher cosine similarity with target tokens and column 3 shows SVMCos's predictions have a lower cosine similarity with non-target tokens (these are true negatives near to the query point). These charts visualise results from a random sample of VSMs evaluated on the BATS dataset.

## 5 Conclusion

We have shown how to fit a max-margin hyperplane to separate source tokens from their related target tokens and that this hyperplane can be used to infer directions in word embedding spaces which correspond to linguistic regularities. This method outperformed all other previous methods in predicting related tokens in a wide variety of vector space models. We have also shown that the offsets between related tokens in embedding spaces are irregular and that several other approaches to predicting them are not as performant.